\documentclass[runningheads]{llncs}
\usepackage{graphicx}
\usepackage{amsmath,amssymb} 
\usepackage{color}
\usepackage[width=122mm,left=12mm,paperwidth=146mm,height=193mm,top=12mm,paperheight=217mm]{geometry}
\usepackage{hyperref}

\usepackage{gensymb}

\definecolor{dgreen}{rgb}{0,.7,0}
\definecolor{dyellow}{rgb}{.7,.7,0}
\definecolor{dred}{rgb}{.7,0,0}
\definecolor{dblue}{rgb}{0,0,0.7}
\definecolor{alexey}{rgb}{0.7,0,1}

\newcommand{\weights}{\psi}
\newcommand{\discr}{D_\varphi}
\newcommand{\gen}{G_\weights}
\newcommand{\inp}{\mathbf{x}}
\newcommand{\targ}{\mathbf{y}}

\begin{document}
\pagestyle{headings}
\mainmatter
\def\ECCV16SubNumber{1592}  

\title{Multi-view 3D Models from Single Images with a Convolutional Network} 

\titlerunning{Multi-view 3D Models from Single Images with a Convolutional Network}

\authorrunning{M. Tatarchenko, A. Dosovitskiy, T. Brox}

\author{Maxim Tatarchenko, Alexey Dosovitskiy, Thomas Brox}
\institute{Department of Computer Science \\ University of Freiburg \\ \texttt{\{tatarchm, dosovits, brox\}{@}cs.uni-freiburg.de}}

\maketitle

\begin{abstract}
We present a convolutional network capable of inferring a 3D representation of a previously unseen object given a single image of this object.
Concretely, the network can predict an RGB image and a depth map of the object as seen from an arbitrary view.
Several of these depth maps fused together give a full point cloud of the object.
The point cloud can in turn be transformed into a surface mesh.
The network is trained on renderings of synthetic 3D models of cars and chairs.
It successfully deals with objects on cluttered background and generates reasonable predictions for real images of cars.
\end{abstract}

\section{Introduction}

The ability to infer a 3D model of an object from a single image is necessary for human-level scene understanding.
Despite the large success of deep learning in computer vision and the diversity of tasks being approached, 3D representations are not yet in the focus of deep networks.  
Can we make deep networks learn such 3D representations? 


\begin{figure}[t]
\begin{center}
   \includegraphics[width=\linewidth]{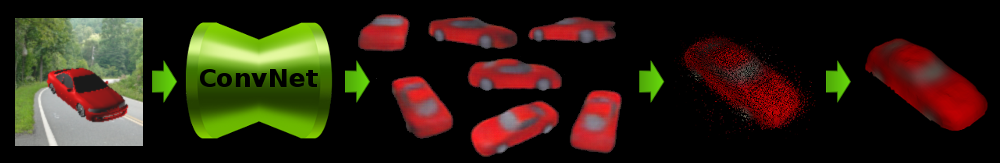}
   \caption{Our network infers an object's 3D representation from a single input image. It then predicts unseen views of this object and their depth maps. Multiple such views are fused into a full 3D point cloud, which is further optimized to obtain a mesh.}
   \label{fig:teaser}
\end{center}
\end{figure}

In this paper, we present a simple and elegant encoder-decoder network that infers a 3D model of an object from a single image of this object, see Figure~\ref{fig:teaser}. 
We represent the object by what we call "multi-view 3D model"~-- the set of all its views and corresponding depth maps.
Given an arbitrary viewpoint, the network we propose generates an RGB image of the object and the depth map.
This representation contains rich information about the 3D geometry of the object, but allows for more efficient implementation than voxel-based 3D models.
By fusing several views from our multi-view representation we get a full 3D point cloud of the object, including parts invisible in the original input image. 

While technically the task comes with many ambiguities, humans are known to be good in using their prior knowledge about similar objects to guess the missing information.
The same is achieved by the proposed network:
when the input image does not allow the network to infer the parts of an object -- for example, because the input only shows the front view of a car and there is no information about its back -- it fantasizes the most probable shape consistent with the presented data (for example, a standard sedan car).

The network is trained end-to-end on renderings of 3D models from the ShapeNet dataset~\cite{savva2015semgeo}.
We render images on the fly during network training, with random viewpoints and lighting.
This makes the training set very diverse, thanks to the size of ShapeNet, and effectively infinite.
We make the task more challenging and realistic by pasting the object renderings on top of random background images.
In this setup, the network learns to automatically segment out the object.
Moreover, we show that networks trained on synthetic images of this kind yield reasonable predictions for real-world images without any additional adaptation.


\textbf{Contributions.} First, we largely improve on the visual quality of the generated images compared to previous work. Second, we achieve this with a simpler and thus more elegant architecture. Finally, we are the first who can apply the network to images with non-homogeneous background and natural images.

\section{Related work}

\paragraph{Unseen view prediction}
Our work is related to research on modeling image transformations with neural-network-based approaches.
These often involve multiplicative interactions, for example gated RBMs~\cite{Memisevic07}, gated autoencoder~\cite{Michalski_NIPS2014} or Disentangling Boltzmann Machines~\cite{Reed_icml2014}.
These approaches typically do not scale to large images, although they potentially could by making use of architectures similar to convolutional DBNs~\cite{Lee_ICML2009}.
They are also typically only applicable to small transformations.

Transforming autoencoders~\cite{HintonKW11} are trained to generate a transformed version of an input image given the desired transformation.
When applied to the NORB dataset of $96 \times 96$ pixel stereo image pairs of objects, this approach can apply small rotations to the input image.

The multi-view perceptron~\cite{ZhuLWT14} is a network that takes a face image and a random vector as input and generates a random view of this face together with the corresponding viewpoint.
In contrast, our model can generate directly the desired view without the need for random sampling.
Kulkarni et al.~\cite{KulkarniWKT15} trained a variant of a variational autoencoder with factored hidden representations, where certain dimensions are constrained to correspond to specific factors of variations in the input data, such as viewpoint and lighting.
This method is conceptually interesting and it allows to generate previously unseen views of objects, but the quality of predictions made by our network is significantly better, as we show in the experimental section.

A simplified version of unseen view prediction is predicting HOG descriptors~\cite{DalalTriggs05} instead of images.
Chen et al.~\cite{ChenG14} pose the problem as tensor completion.
Su et al.~\cite{Su_ICCV2015} find object parts similar to those of a given object in a large dataset of 3D models and interpolate between the desired views of these. 
These methods do not learn a 3D representation of the object class but approximate unseen views by linear combinations of models from a fixed dataset.

Dosovitskiy et al.~\cite{DB15} trained an 'up-convolutional' network to generate an image of a chair given the chair type and a viewpoint.
This method is restricted to generating images of objects from the training set or interpolating between them.
Applying the method to a new test image requires re-training the network, which takes several days.
While the decoder part of our network is similar to the architecture of Dosovitskiy et al., our network also includes an encoder part which infers the high-level representation from a given input image.
Hence, at test time we can generate unseen views and depth maps of new objects by simply forward propagating an image of the object through the network.
Our approach also yields more accurate predictions.

Most closely related is the concurrent work by Yang et al.~\cite{yang15,yang15_arxiv}. 
They train a recurrent network that can rotate the object in the input image: given an image, it generates a view from a viewpoint differing by a fixed increment.
This makes the approach restricted to generating a discrete set of views, while we are able to vary the angle continuously.
In the approach of Yang et al., one might train the network with a small angle increment and predict views at finer quantization levels than the $15$ degrees used by the authors. However, this would require more recurrent iterations for performing large rotations. It would be slow and probably would lead to error accumulation. Our network does not have such restrictions and produces an arbitrary output view in a single forward pass. 
Moreover, it can generate a full 3D point cloud, can deal with non-homogeneous background, and the generated images are of much better quality. 

\paragraph{3D from single image}
Inferring a 3D model of an object from a single image is a long-standing, very difficult task in computer vision.
A general approach is to use certain models of lighting, reflectance and object properties to disentangle these factors given a 2D input image~\cite{Barron_TPAMI2015}.
When reconstructing a specific object class, prior knowledge can be exploited.
For example, morphable 3D models~\cite{Blanz_PAMI2003}, \cite{LiuZLZ15} are commonly used for faces.
Kar et al.~\cite{KTCM15} extended this concept to object categories with more variation, such as cars and chairs, and combined it with shape-from-shading to retrieve also the high frequency components of the shape.
For building their morphable 3D model they rely on ideas from Vicente et al.~\cite{VCAB14}, who showed that the coarse 3D structure can be reconstructed from multiple images of the same object class (but different object instances) and some keypoint annotation.
In contrast to Kar et al.~\cite{KTCM15}, our approach does not use an explicit 3D model.  A 3D model representation for the object class is rather implicit in the weights of the convolutional network.

Aubry et al. \cite{Aubry14} proposed an approach for aligning 3D models of objects with images of these objects.
The method makes use of discriminative part detectors and works on complicated real scenes.
On the downside, this is a nearest-neighbor kind of method: it selects the best fitting 3D models from a fixed set of models.
This limits the generalization capability of the method and makes it proportionally slower if the model collection grows in size.

Huang et al.~\cite{HuangWK15} reconstruct 3D models from single images of objects by jointly analyzing large collections of images and 3D models of objects of the same kind.
The method yields impressive results. However, it jointly processes large collections of images and models with a nearest neighbor approach and hence cannot be applied to a new image at test time that is different from all models in the dataset. 

Eigen et al.~\cite{Eigen_NIPS2014} trained convolutional networks to predict depth from single images of indoor scenes.
This is very different from our work in that we predict depth maps not only for the current viewpoint, but also for all other viewpoints.
Wu et al.~\cite{Wu_CVPR2015} trained 3D Convolutional Deep Belief Networks capable of generating a volumetric representation of an object from a single depth map.
This method requires a depth map as input, while our networks only take a single RGB image.

\section{Model description}

We train a network that receives an input pair $(x_i, \theta_i)$, where $x_i$ is the input image and $\theta_i$ the desired viewpoint, and aims to estimate a pair $(y_i, d_i)$, where $y_i$ is the 2D projection of the same object from the requested viewpoint and $d_i$ is the depth map of this projection.
While the input images $x_i$ may have complicated background, the targets $y_i$ always have monotonous background.
$\theta_i$ is a vector defining the viewpoint; it consists of two angles~-- azimuth $\theta_i^{az}$ and elevation $\theta_i^{el}$~-- and the distance $r$ from the object center.
Angles are given by their sine and cosine to deal with periodicity.
The viewpoint of the input image is \emph{not} given to the network. 
This makes the task more difficult since the network must implicitly infer the viewpoint from the input image.

The network is trained by minimizing the loss function $\mathcal{L}$ which is a weighted sum of two terms: squared Euclidean loss for the RGB image and $L_1$ loss for the depth image:
\begin{equation}
\mathcal{L} = \sum\limits_{i} ||y_i - \widehat{y}_i||_2^2 + \lambda\, ||d_i - \widehat{d}_i||_1,
\end{equation}
where $\widehat{y}_i$ and $\widehat{d}_i$ are the outputs of the network and $\lambda$ is the weighting coefficient.
We used $\lambda=0.1$ in our experiments.

\subsection{Architecture}

\begin{figure*}
\begin{center}
\includegraphics[width=\linewidth]{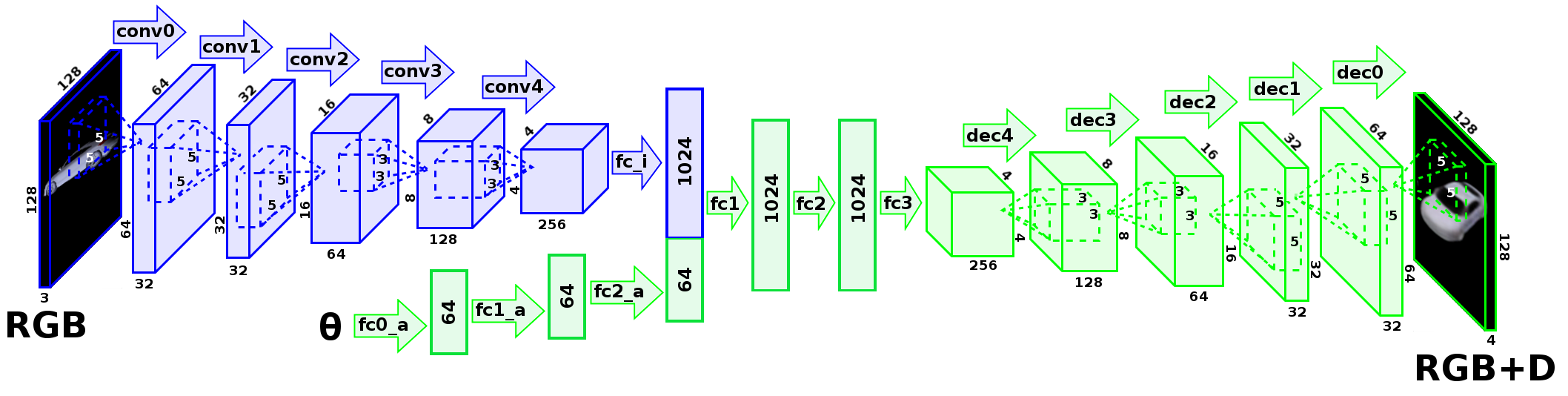}
\end{center}
   \caption{The architecture of our network.
   The encoder (\textbf{blue}) turns an input image into an abstract 3D representation. The decoder (\textbf{green}) processes the angle, modifies the encoded hidden representation accordingly, and renders the final image together with the depth map.}
\label{fig:architecture}
\end{figure*}

The architecture of our encoder-decoder network is shown in Figure~\ref{fig:architecture}. 
It is simple and elegant. The encoder part (blue in the figure) processes the input image to obtain a hidden 3D representation $z_{obj}$ of an object shown in the image.
The decoder part (green in the figure) then takes $z_{obj}$ and the desired viewpoint as inputs and renders the final output image.

During training, the network is always presented with pairs of images showing two views of the same object together with the viewpoint of the output view.
Objects are randomly sampled from a large database of 3D models, and pairs of views are randomly selected.

Technically, the encoder part propagates an input image through a standard ConvNet architecture, which consists of 5 convolutional layers with stride $s=2$ in each layer and one fully connected layer in the end.
The decoder part independently processes the angle in $3$ fully connected (FC) layers, then merges the resulting code with the output of the encoder and performs joint processing in 3 more FC layers.
Finally, it renders the desired picture using $5$ up-convolutional layers (also known as "deconvolutional").
We experimented with deeper and wider networks, but did not observe a significant difference in performance.

The up-convolutional layers perform upsampling+convolution, opposite to the standard convolution+pooling. During upsampling, each pixel is replaced with a $2 \times 2$ block containing the original pixel value in the top left corner and zeros {everywhere} else.
For both convolutional and up-convolutional layers of the network we use $5 \times 5$ filters for outer layers and $3 \times 3$ filters for deeper layers.

The Leaky ReLU nonlinearity with the negative slope $0.2$ is used after all layers, except for the last one, which is followed by the $tanh$.

\subsection{Multi-view 3D to point cloud and mesh}

The multi-view 3D model provided by the network allows us to generate a point cloud representing the object, which in turn can be transformed into a mesh.
To achieve this, for a single input we generate multiple output images from different viewpoints together with their corresponding depth maps.
The camera parameters are known: both internal (focal length, camera model) and external (camera pose). 
This allows us to reproject each depth map to a common 3D space and obtain a single point cloud.

As a post-processing step we can turn the point cloud into a dense surface model with the method of Pock et al.~\cite{PockZB11}.
This method uses depth information together with the point normals to compute the final mesh.
As the normal information is missing in our case (although it potentially could also be estimated by the network), we approximate it by providing the direction to the camera for each point.
Since the normals are optimized anyway by the fusion method, this approximation yields good results in practice.

\begin{figure}[t]
\begin{center}
   \includegraphics[width=\linewidth]{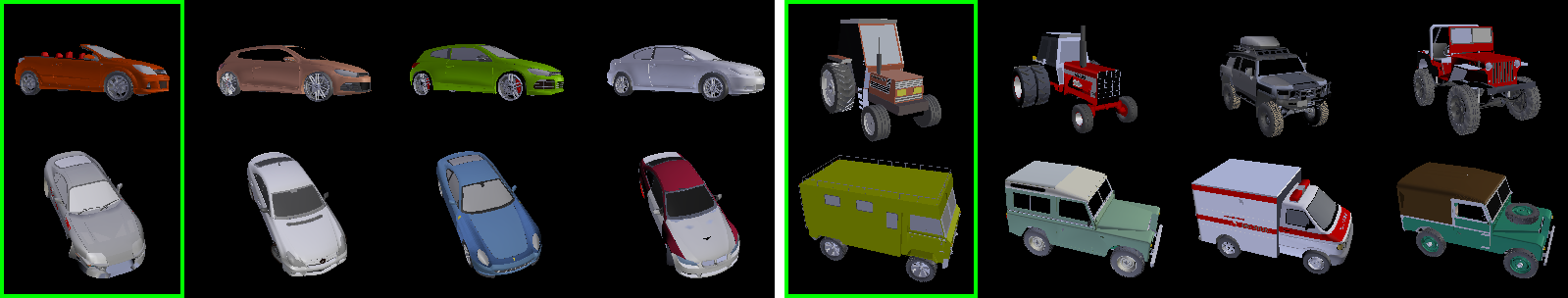}
   \caption{Train-test split of cars. Sample renderings and their nearest neighbors are shown. Each row shows on the left a rendering of a query model from the test set together with several HOG space nearest neighbors from the training set. The two query models on the right are 'difficult' ones.
   \label{fig:example_renders}}
\end{center}
\end{figure}

\subsection{Dataset}

We used synthetic data from the ShapeNet dataset \cite{savva2015semgeo} for training the networks.
The dataset contains a large number of 3D models of objects belonging to different classes.
The models have been semi-automatically aligned to a consistent orientation using a hierarchical approach based on \cite{Huang_2013}.
We mainly concentrated on car models, but we also trained a network on chairs to show generality of our approach and to allow a comparison to related methods.
We used $7039$ car models and $6742$ chair models.


3D models were rendered using our self-developed real-time rendering framework based on the Panda3D rendering engine\footnote{\url{https://www.panda3d.org}}.
This allowed us to generate training images on the fly, without the need to store huge amounts of training data on the hard drive.
We randomly sampled azimuth angles in the range from $0\degree$ to $360\degree$, elevation angles in the range from $-10\degree$ to $40\degree$, and the distance to the object from 1.7 to 2.3 units, with a car length being approximately equal to 3 units.

We took special care to ensure the realism of the renderings, since we would like the network to generalize to real input images.
As in Su et al.~\cite{Su15}, we randomly sampled the number of light sources from 2 to 4, each with random intensity and at random location.
When overlaying the rendering on top of the background, we performed alpha compositioning to avoid sharp transition.
It was implemented by smoothing the segmentation mask with a Gaussian filter with the standard deviation randomly sampled between $1$ and $1.3$.
Additionally, we smoothed the car image with a Gaussian filter with the standard deviation randomly sampled between $0.2$ and $0.6$.

Since we used a large amount of models, some of which may happen to be similar, simple random train-test splitting does not enable a reliable evaluation.
To mitigate this problem we clustered objects according to their similarity and then took some of the clusters as the test set.
We are mostly interested in splitting the objects according to their shape, so we used the distance between the 2D HOG descriptors~\cite{DalalTriggs05} of the corresponding images as similarity measure.
To make this measure more robust, we considered three different viewpoints for each object and used the sum of three distances as the final distance measure.
After constructing a matrix of pairwise distances, we clustered the models using agglomerative clustering with average linkage.

For cars we selected a single cluster consisting of 127 models as the test set.
Models from this group we refer to as 'normal test cars'.
In addition, we picked 20 more models from the training set that have the highest distance from 'normal cars' and added them to the test set.
Those are referred to as 'difficult test cars'.
Example models from the test set and their corresponding nearest neighbors from the training set are shown in Figure~\ref{fig:example_renders}.
For chairs we picked three clusters as the test set comprising a total of 136 models.

\section{Experimental evaluation}
  
\paragraph{Network training details}
We used Tensorflow~\cite{tensorflow2015-whitepaper} for training the networks.
The objective was optimized using the Adam method \cite{KingmaB15} with $\beta_1=0.9$ and $\beta_2=0.999$.
We initialized the weights of our network by sampling a Gaussian with corrected variance as described in \cite{HeZRS15}.
The learning rate was equal to $0.0001$.

We did not perform data augmentation, as we observed that it does not result in better generalization but leads to slower convergence.
It seems there is already enough variation in the training data.

\subsection{Unseen view prediction}

\begin{figure}[t]
\begin{center}
   \includegraphics[width=\linewidth]{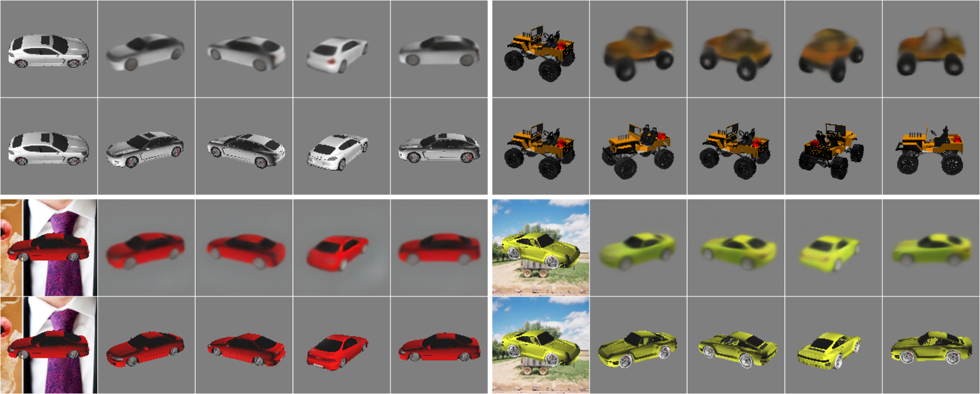}
   \caption{Predictions of the network (top row for each model) and the corresponding ground truth images (bottom row for each model).
   The input to the network is in the leftmost column for each model.
   The top right model is a "difficult" car.\label{fig:example_random_cars}}
   \end{center}
\end{figure}

We trained the networks to generate previously unseen views of objects from a single input image, therefore this is the first task we test on.
Exemplary results for cars are shown in Figure~\ref{fig:example_random_cars}.
The network predicts the desired view for both normal and difficult (top right) cars, without (top row) and with (bottom row) background.
The shape and the color of the car are always correctly estimated.
Predictions for the difficult car are more blurry, since this car is dissimilar from models the network has been trained on.

Compared to the ground truth, the predictions are slightly blurry and lack some details. Apart from the fact that the problem is heavily ill-posed, this is likely to be a consequence of using squared Euclidean error as the loss function: if the network is uncertain about the prediction, it averages over potential images, resulting in blur.
This could be mitigated for example by adversarial training, first proposed by Goodfellow et al.~\cite{GoodfellowPMXWOCB14}. 
We experimented in this direction (see supplementary material for details), and indeed the images become slightly sharper, but at the cost of introduced noise, artifacts and very sensitive training.
Therefore, we stick with the squared Euclidean loss. 

\begin{figure}[t]
\begin{center}
   \includegraphics[width=\linewidth]{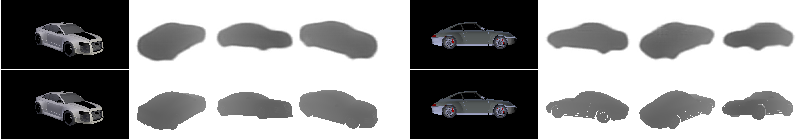}
   \caption{Depth map predictions (\textbf{top row}) and the corresponding ground truth (\textbf{bottom row}).
   The network correctly estimates the shape.}
   \label{fig:depth_maps}
   \end{center}
\end{figure}

\begin{table}
\small
\begin{center}
\begin{tabular}{ | l | c | c | c | c | }
  \hline
   & \multicolumn{2}{c|}{Color} & \multicolumn{2}{c|}{Depth}\\
   \hline
   & Normal & Difficult & Normal & Difficult \\
  \hline
  NN HOG & 0.028 & 0.039 & 0.0058 & 0.0225 \\
  \hline
  NN HOG+RGB & 0.020  & 0.036 & 0.0058  & 0.0221 \\
  \hline
  NN RGB & 0.018  & 0.034 & 0.0064  & 0.0265 \\
  \hline
  Network & \textbf{0.013}  & \textbf{0.028} & \textbf{0.0057}  & \textbf{0.0207} \\
  \hline
\end{tabular}
\end{center}
\caption{Average error of predicted unseen views with our network and with the nearest neighbor baseline.}
\label{tbl:rgb_baseline}
\end{table}


\subsubsection{Comparison with a nearest neighbor baseline}

We compare the network with a simple nearest neighbor (NN) baseline approach.
We maximally simplify the task for the baseline approach: unlike the network it knows the input image viewpoint and there is no background.
Given an input image with known viewpoint, the baseline searches the training set for the model which looks most similar from this viewpoint according to some metric.
The prediction is simply the rendering of this model from the desired viewpoint. 
We tried three different metrics for the NN search: Euclidean distance in RGB space, Euclidean distance in HOG space, and a weighted combination of these.

Table~\ref{tbl:rgb_baseline} reports average errors between the ground truth images and the predictions generated either with the baseline method or with our network.
The error measure is Euclidean distance between the pixel values, averaged over the number of pixels in the image, the number of input and output viewpoints, the number of models and the maximum per pixel distance (443.4 for RGB and 65535 for depth).
We separately show results for normal and difficult cars.

The network outperforms the baselines on both tasks, even though it is not given the input viewpoint.
NN search can yield cars that look alike from the input view but may be very different when viewed from another angle.
The network, in contrast, learns to find subtle cues which help to infer the 3D model.
Another clear disadvantage of the NN search is that it can only return what is in the dataset, whereas the network can recombine the information of the training set to create new images.


\subsubsection{Comparison with existing work}

We compared our results to several existing deep learning approaches that generate unseen views of images.

Except for a comparison to Dosovitskiy et al.~\cite{DB15}, for which code was available, all comparisons are only on a qualitative basis. There are two reasons: first, there was no code to run other existing methods. Second, it is unclear which quantitative measure would be best to judge the quality of generated images. The best quantitative experiment would be a study with human observers, who have to assess which images look better. Since the differences in the quality of the results is mostly so obvious that quantitative numbers would not provide additional information, the lack of code is not a problem. 

\begin{figure}
\centering
\begin{minipage}{.45\textwidth}
  \begin{center}
   \includegraphics[width=\linewidth]{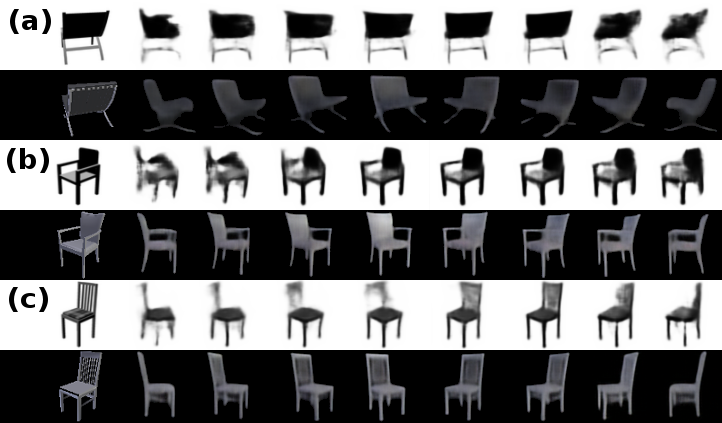}
   \caption{Our results (black background) compared with those from IGN \cite{KulkarniWKT15} (white background) on similar chair models.
   The leftmost image in each row is the input to the network.
   In all cases our results are much better.}
   \label{fig:ign_comparison}
   \end{center}
\end{minipage}%
\hspace{0.3cm}
\begin{minipage}{.08\textwidth}
\end{minipage}%
\begin{minipage}{.45\textwidth}
  \begin{center}
   \includegraphics[width=\linewidth]{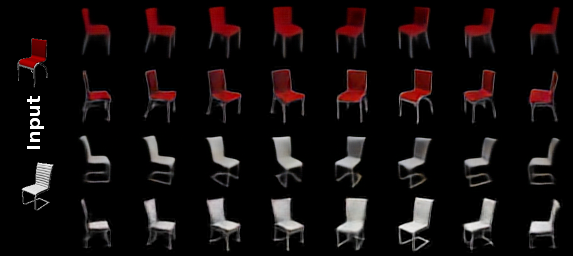}
   \caption{Comparison of our approach (top row for each model) with novel view prediction results from Dosovitskiy et al.~\cite{DB15} (bottom row for each model). The estimates of our network are more accurate and consistent, and it does not require re-training for each new model.}
   \label{fig:chairs_comparison}
   \end{center}
\end{minipage}
\end{figure}

In order to compare with the Inverse Graphics Network (IGN) of Kulkarni et al. \cite{KulkarniWKT15} we selected from our test set chair models similar to those Kulkarni et al. used for testing and showed in their paper.
We also used the same input viewpoint.
The results are shown in Figure~\ref{fig:ign_comparison}.
In all cases our network generates much more accurate predictions.
Unlike IGN, it always predicts the correct view and generates visually realistic and detailed images.
It captures fine details like bent legs in the top example or armrests in the second example.

We also compared to Dosovitskiy et al.~\cite{DB15}.
This approach allows the prediction of all views of a chair model given only a single view during training.
However, 1) it requires several days of training to be able to predict unseen views of a new chair and 2) it is not explicitly trained to predict these unseen views, so there is no guarantee that the predictions would be good.
We used the code provided by the authors to perform comparisons shown in Figure~\ref{fig:chairs_comparison}.
For each model the top row shows our predictions and the bottom row those from Dosovitskiy et al.
While in simple cases the results look qualitatively similar (top example), our approach better models the details of the chair style (chair legs in the bottom example).
This is supported by the numbers: the average error of the images predicted by our network is $0.0208$ on the chairs dataset, whereas the network of Dosovitskiy et al. has an average error of $0.0308$.
The error measure is the same as in the baseline comparison.

\begin{figure}[t]
\begin{center}
   \includegraphics[width=\linewidth]{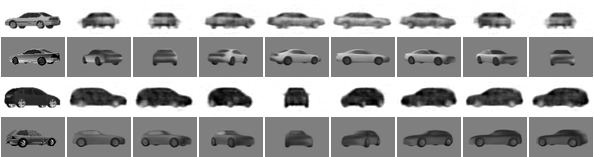}
   \caption{Predictions from Yang et al. \cite{yang15_arxiv} (top row for each model) compared to our predictions (bottom row for each model) on similar car models. The leftmost image in each row is the input to the network. Our network generates more realistic and less blurred images.}
   \label{fig:yang}
   \end{center}
\end{figure}

Finally, we qualitatively compared our results with the recent work of Yang et al. \cite{yang15_arxiv}.
Here we show the results on cars, which we found to be more challenging than chairs. 
Figure~\ref{fig:yang} shows predictions by Yang et al. (top row for each model) and our work (bottom row for each model). 
For both models the leftmost column shows the input image. We picked the cars from our dataset that most resemble the cars depicted in Yang et al.~\cite{yang15_arxiv}.
Since images generated by their method are $64 \times 64$ pixels, we downsampled our results to this resolution to make the visual comparison fair. 
Our predictions look much more realistic and significantly less blurred.
The method of Yang et al. occasionally averages several viewpoints (for example, the third column from the left for the top model in Figure~\ref{fig:yang}), while our method always generates sharp images as seen from the desired viewpoint.

\subsubsection{Natural input images}

\begin{figure}[t]
\begin{center}
   \includegraphics[width=\linewidth]{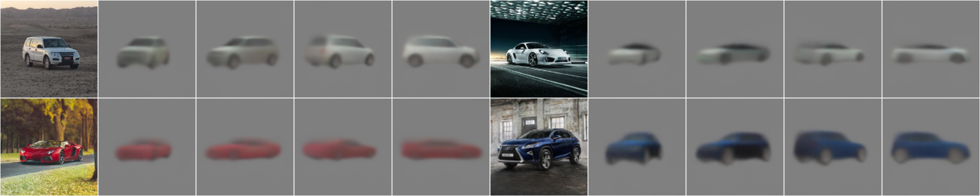}
   \caption{Network predictions for natural input images. The net correctly estimates the shape and color.}
   \label{fig:real_cars}
   \end{center}
\end{figure}

To verify the generalization properties of our network, we fed it with images of real cars downsampled to the size of $128 \times 128$ pixels.
The results are shown in Figure~\ref{fig:real_cars}\,.
We do not have ground truth for these images so only the output of the network is shown.
The quality of the predictions is slightly worse than for the (synthetic) test cars.
The reasons may be complicated reflections and camera models different from ones we used for rendering. 
Still, the network estimates the shape and the color well.

We observed that the realistic rendering procedure we implemented is important for generalization to real images.
We show in the supplementary material that simpler rendering leads to complete failure on real data.

We emphasize that this is the first time neural networks are shown to be able to infer 3D representations of objects from real images.
Interesting avenues of future research include deeper study of the network's performance on real images, as well as joint training on synthetic and real data and applications of transfer learning techniques.
However, these are beyond the scope of this paper.


\subsection{3D model prediction}

\begin{figure}[]
\begin{center}
   \includegraphics[width=\linewidth]{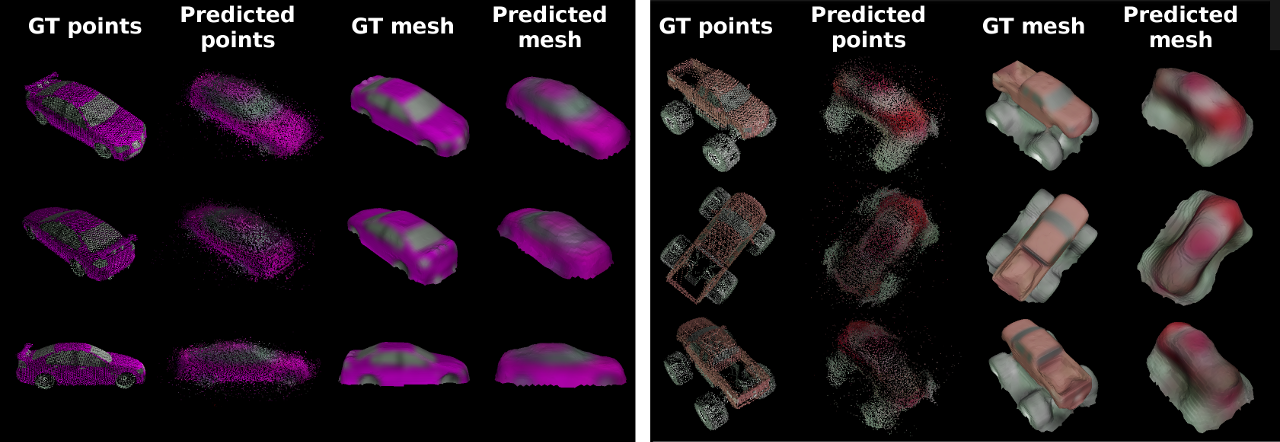}
   \caption{3D model reconstructions of a "normal" (left) and a "difficult" (right) car.}
   \label{fig:example_3d_models}
   \end{center}
\end{figure}

We verified to which extent the predicted depth maps can be used to reconstruct full 3D surfaces.
Figure~\ref{fig:depth_maps} shows two exemplary depth maps generated by our network together with the corresponding ground truth.
The overall quality is similar to that of predicted images: the shape is captured correctly while some fine details are missing.


In Figure~\ref{fig:example_3d_models} we show 3D models obtained by fusing 6 predicted depth maps ($\theta^{el}=20\degree$, $\theta^{az}=\{0\degree,60\degree,120\degree,180\degree,240\degree,300\degree\}$).
Already the raw point clouds represent the shape and the color well, even for the "difficult" model (right).
Dense depth map fusion removes the noise and a smooth surfaces, yet also destroys some more details due to the regularizer involved. 
For more results on 3D models we refer to the supplementary video \url{https://youtu.be/uf4-l6h7iGM}.



\subsection{Analysis of the network}


\subsubsection{Viewpoint dependency}

\begin{figure}
\centering
\begin{minipage}{.45\textwidth}
  \begin{center}
   \includegraphics[width=\linewidth]{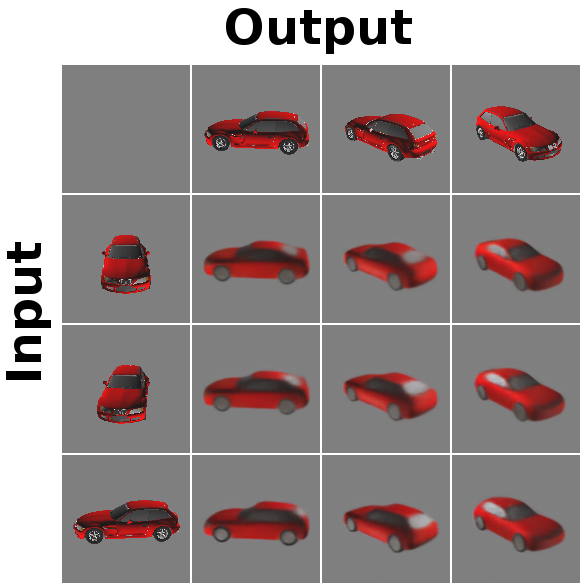}
   \caption{The more informative the input view is, the better the network can estimate the ground truth image. For uninformative inputs it simply invents some model which is still internally consistent.}
   \label{fig:viewpoint_dependency}
   \end{center}
\end{minipage}%
\hspace{0.3cm}
\begin{minipage}{.08\textwidth}
\end{minipage}%
\begin{minipage}{.45\textwidth}
  \begin{center}
   \includegraphics[width=\linewidth]{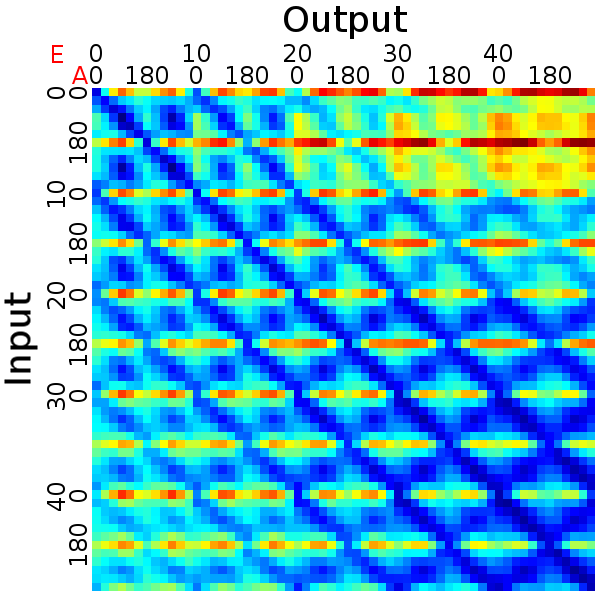}
   \caption{Distance from ground truth for different input and output views. Shown are all combinations of elevation (\textbf{E}) and azimuth (\textbf{A}) angles with a $30\degree$ step. It is harder to predict views that are significantly different from the input.}
   \label{fig:confusion_matrix}
   \end{center}
\end{minipage}
\end{figure}

Since the prediction task is ambiguous, the quality of predictions depends on how informative the input image is with regard to the desired output view.
For our network we can observe this tendency, as shown in Figure~\ref{fig:viewpoint_dependency}\,.
If the input viewpoint reveals much about the shape of the car, such as the side-view input, the generated images match the ground truth quite well.
In case of less informative input, such as the front-view input, the network has to do more guesswork and resorts to predicting the most probable answer.
However, even if the input image is weakly informative, all the predicted views correspond to a consistent 3D shape, indicating that the network first extracts a 3D representation from the image and then renders it from different viewpoints.


In Figure~\ref{fig:confusion_matrix} we quantify the prediction quality depending on the input and output views.
The matrix shows the Euclidean distance between the generated and ground truth images for different input (y-axis) and output views (x-axis) averaged over the whole test set.
Each column is normalized by its sum to compensate for different numbers of object pixels in different views.
Several interesting patterns can be observed.
The prediction task gets harder if the input view is very different from the output view, especially if the input elevation is small: top right and bottom left corners of the matrix are higher than the rest.
Local patterns show that for each elevation it is easier to predict images with the same or similar azimuth angles.
Diagonal blue stripes show that it is easier to predict similar or symmetric views.


\subsubsection{Object interpolation}

\begin{figure}[t]
\begin{center}
   \includegraphics[width=\linewidth]{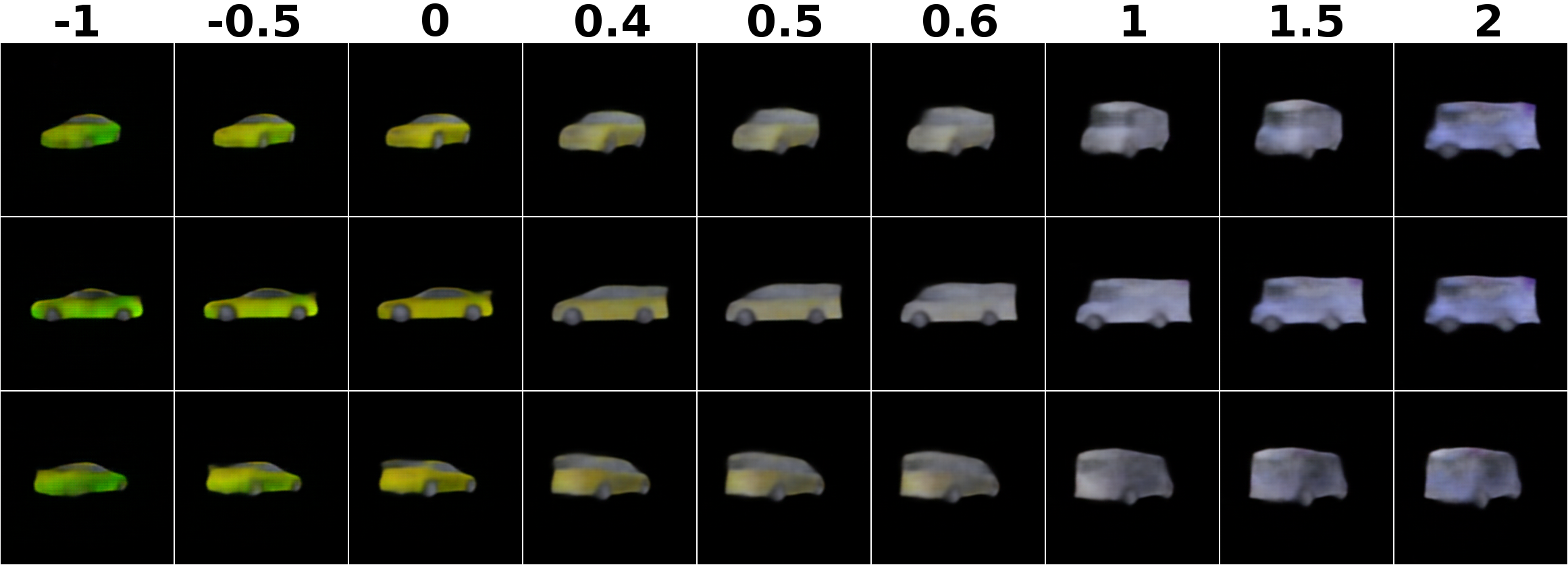}
   \caption{Morphing a car into a bus by interpolating between the feature representations of those two models.
   All the intermediate models are consistent.}
   \label{fig:interpolation_models}
   \end{center}
\end{figure}

The hidden object representation extracted by the network is not directly interpretable.
One way to understand it is to modify it and see how this affects the generated image.
In the experiment shown in Figure~\ref{fig:interpolation_models}, we encoded two extremely different models (a car and a bus) into feature vectors $f_{car}$ and $f_{bus}$, linearly interpolated between these $f_{int} = \alpha f_{car} + (1-\alpha) f_{bus}$, and decoded the resulting feature vectors.
We also tried extrapolation, that is, $\alpha < 0$ and $\alpha > 1$.

The first and most important observation is that all generated views form consistent shapes, which strongly indicates that the interpolation modifies the 3D representation, which is then rendered from the desired viewpoint.
Second, extrapolation also works well, exaggerating the 'carness' or the 'busness' of the models.
Third, we observed that the morphing is not uniform: there is not much happening for $\alpha$ values close to $0$ and $1$, most of the changes can be seen when $\alpha$ is around $0.5$.

\begin{figure*}[t]
\begin{center}
\includegraphics[width=\linewidth]{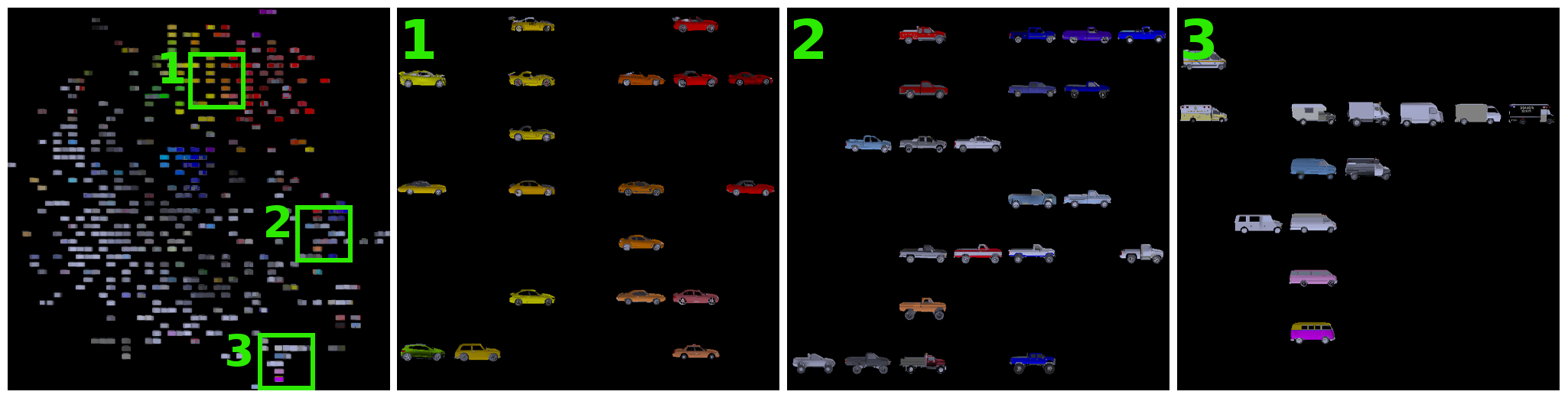}
\caption{t-SNE embedding in latent 1024-dimensional space. Cars are grouped according to their shape and color.}
\label{fig:tsne}
\end{center}
\end{figure*}

\subsubsection{Internal representation}



In order to study the properties of the internal representation of the network, we ran the t-SNE embedding algorithm \cite{ictdbid:2777} on the 1024-dimensional vectors computed for a random subset of models from the training set with fixed viewpoint.
t-SNE projects high-dimensional samples to a 2D space such that similar samples are placed close to one another.
The results of this experiment are shown in Figure~\ref{fig:tsne}.
Both shape and color are important, but shape seems to have more weight: similar shapes end up close in the 2D space and are sorted by color within the resulting groups.

In Section 3 of the supplementary material we also show that different input views of the same object lead to very similar intermediate representations.

\section{Conclusions}

We have presented a feed-forward network that learns implicit 3D representations when being trained on the task to generate new views from a single input image.
Apart from rendering any desired view of an object, the network allows us to also generate a point cloud and a surface mesh.
Although the network was trained only on synthetic data, it can also take natural images as input. 
Clearly, natural images are harder for the network since the training data does not yet fully model all variations that appear in such images. In future work we will investigate ways to improve the training set either by more realistic renderings or by ways to mix in real images.

\section{Acknowledgments}

We  acknowledge  funding  by  the  ERC  Starting  Grant VideoLearn  (279401).   We  would  like  to thank Nikolaus Mayer and Benjamin Ummenhofer for their comments.


\bibliographystyle{splncs}
\bibliography{carsbibliography}

\clearpage

\section*{Supplementary Material}
\renewcommand{\thesubsection}{\Alph{subsection}}

We present experimental results showing the effect of realistic rendering and the effect of adversarial training. We also analyze how the internal representation changes when the network is presented with different input views of the same object.

\subsection{Realistic rendering}

As mentioned in the paper, we found that in order to achieve better generalization to real images special care has to be taken when rendering the training data.
We trained networks with two kinds of training data: "realistic" and "basic".

The "realistic" rendering is described in the main paper: we randomly sampled the number of light sources, their intensities and the locations; performed alpha compositioning to avoid sharp transition between the model and the background; and additionally smoothed the car image with a Gaussian filter.
The "basic" rendering is with two light sources of fixed intensity, without alpha compositioning and smoothing.

\begin{figure*}
\begin{center}
\includegraphics[width=\linewidth]{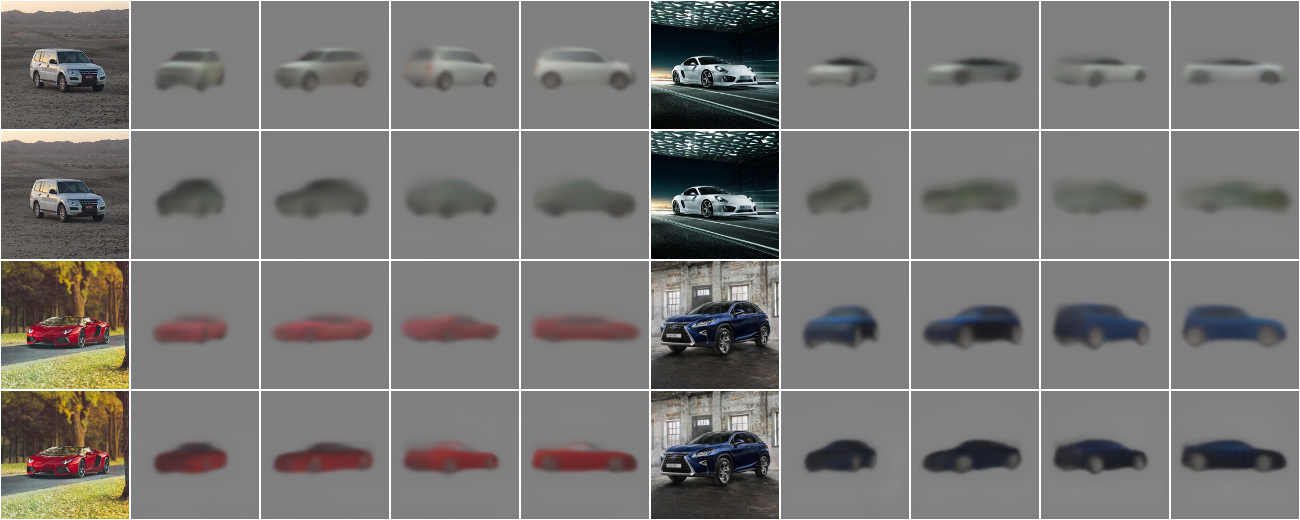}
\end{center}
   \caption{Predictions from the network trained on "realistic" data (top for each model) compared with those from the network trained on "basic" data (bottom for each model).}
\label{fig:realistic_rendering}
\end{figure*}

Figure~\ref{fig:realistic_rendering} compares the results of networks trained on these two kinds of data.
The network trained on "basic" data (bottom row for each model)  fails to correctly estimate the car shape in all cases but one.
The network trained with "realistic" data performs much better, demonstrating how the  quality of the training data is crucial for generalization to real images.

\subsection{Adversarial training}
 
Tasks involving image generation are still mostly solved by optimizing $L_2$ objective, which is robust but often leads to blurred results.
This happens because of the fundamental uncertainty associated with novel view estimation, which in case of Euclidean loss leads to predicting the average of all possibilities.
Alternatively, one can use the idea of adversarial training  introduced by Goodfellow et al. \cite{gan14}.
The aim is to train a \emph{generator} $\gen$ (parametrized by a neural network with weights $\psi$) which takes random noise as input and generates realistic images.
This is achieved by training the generator concurrently with another neural network ~-- a \emph{discriminator} $\discr$. 
The discriminator aims to distinguish the generated images from real ones, while the generator aims to trick the discriminator.
Mathematically, the parameters $\varphi$ of the discriminator are trained by minimizing
\begin{equation} \label{eq:discrloss_discr}
 \mathcal{L}_{discr} = -\sum\limits_{i} \log (\discr(\targ_i)) + \log (1 - \discr(\gen(\inp_i))), 
\end{equation}
where $\inp_i$ is the noise sample and $\targ_i$ is the target sample from the training set.
The generator is trained to minimize
\begin{equation} \label{eq:discrloss_gen}
 \mathcal{L}_{adv} = -\sum\limits_{i} \log \discr(\gen(\inp_i)).
\end{equation}

Conditional GANs were successfully applied to future prediction in videos~\cite{mathieu15} and other image generation tasks~\cite{DosovitskiyB16} and demonstrated superior performance over standard squared Euclidean objective.
This motivated us to use adversarial training to decrease blur in car images predicted by the network.
Namely, we minimized
\begin{equation}
\mathcal{L}_{gen} = \mathcal{L}_{euc} + \alpha \mathcal{L}_{adv},
\end{equation}
where $\mathcal{L}_{adv}$ was trained as described above, with the difference that our generator was conditioned on the input image instead of noise.
In our experiments we used $\alpha=0.01$. 
We used the same generator as for all other experiments.
The discriminator is a convolutional network identical to the encoder of the generator.
It takes both input and output view as input.

\begin{figure*}
\begin{center}
\includegraphics[width=\linewidth]{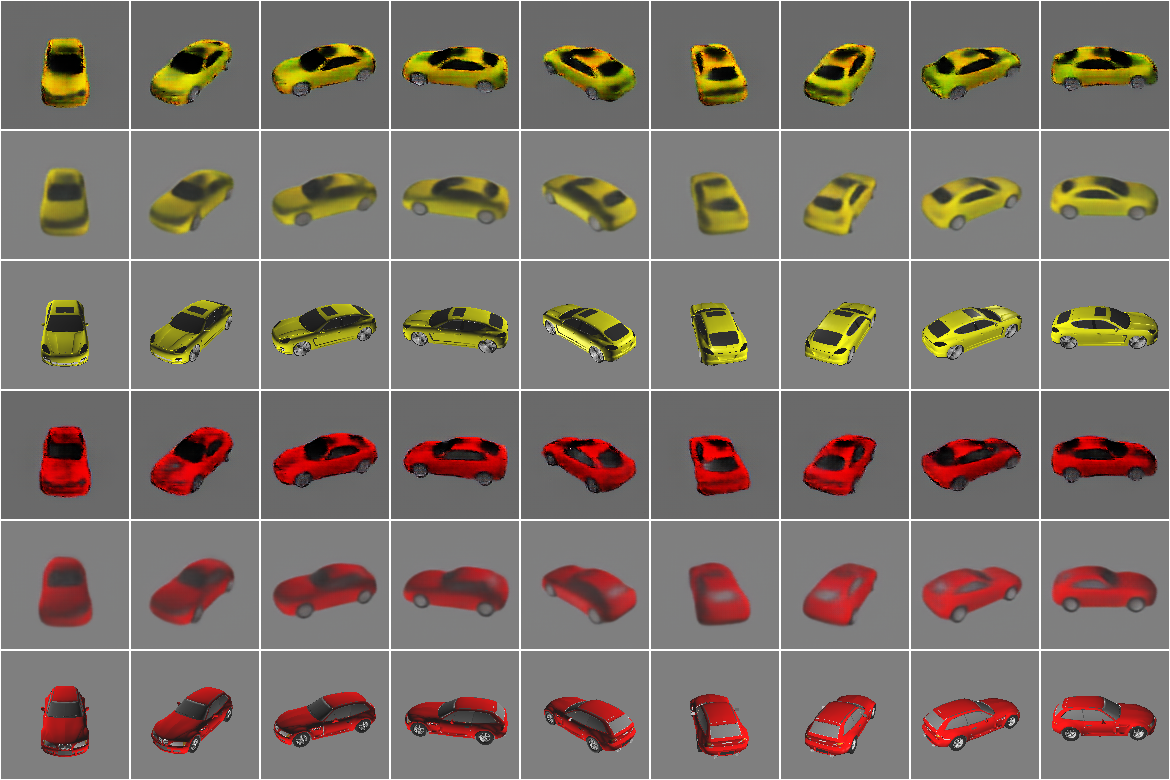}
\end{center}
   \caption{Predictions with networks trained with adversarial and squared Euclidean loss. For each model, top row: with adversarial loss, second row: without adversarial loss, bottom row: ground truth.}
\label{fig:adversarial}
\end{figure*}

Comparison of viewpoint prediction results with and without adversarial loss is shown in Figure~\ref{fig:adversarial}.
While adversarial training does lead to sharper predictions, this happens at the cost of increased image noise and worse estimate of the car shape.
Moreover, the network with adversarial loss is much more sensitive to hyperparameter settings.
We therefore concentrated on getting best results with standard non-adversarial losses. 
Still, we believe adversarial training could be useful to increase the visual quality of the network predictions and see it as an interesting direction of future research.

\subsection{Intermediate representation}

\begin{figure*}
\begin{center}
\includegraphics[width=0.7\linewidth]{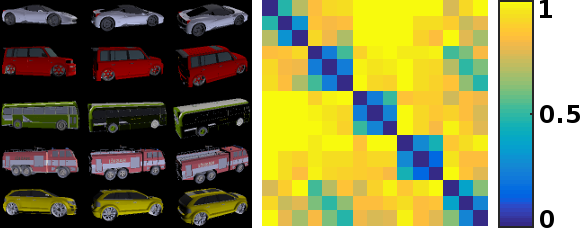}
\end{center}
   \caption{Pairwise distances between the hidden vectors of five cars  and  three  input  views. Different  input  views  of  the  same model lead to similar hidden representations.}
\label{fig:dist_mat}
\end{figure*}

We studied the properties of the internal representation by computing it for 3 different views of 5 different car models. Figure~\ref{fig:dist_mat} shows the input data and the matrix of pairwise Euclidean distance between the cars in the hidden space. $3\times3$ diagonal blocks indicate that different input views of
the same car lead to a similar hidden representation. The representation of the second car is quite close to that of the fifth one (off-diagonal blue elements in the matrix) because both cars have similar shape.

\end{document}